\documentclass[10pt,twocolumn,letterpaper]{article}

\usepackage{iccv}
\usepackage{times}
\usepackage{epsfig}
\usepackage{graphicx}
\usepackage{amsmath}
\usepackage{amssymb}

\usepackage{color}
\usepackage{siunitx}
\usepackage{mathtools}
\usepackage{multirow}
\usepackage{booktabs}
\usepackage{subcaption}
\usepackage{gensymb}
\usepackage{colortbl}
\usepackage{arydshln}
\usepackage{multicol}

\usepackage{amsmath,amsfonts,bm}

\def\eqref#1{equation~\ref{#1}}

\def\1{\bm{1}}

\DeclareMathAlphabet{\mathsfit}{\encodingdefault}{\sfdefault}{m}{sl}
\SetMathAlphabet{\mathsfit}{bold}{\encodingdefault}{\sfdefault}{bx}{n}

\newcommand{\R}{\mathbb{R}}

\usepackage[breaklinks=true,bookmarks=false]{hyperref}

\hypersetup{
    colorlinks=true,
    linkcolor=blue,
    filecolor=cyan,      
    urlcolor=magenta,
}
\iccvfinalcopy 

\def\iccvPaperID{****} 

\ificcvfinal\fi

\begin{document}

\title{CrossNorm and SelfNorm for Generalization under Distribution Shifts}

\author{Zhiqiang Tang\\
Amazon Web Services\\
{\tt\small zqtang@amazon.com}
\and
Yunhe Gao\\
Rutgers University\\
{\tt\small yunhe.gao@rutgers.edu}
\and
Yi Zhu\\
Amazon Web Services \\
{\tt\small yzaws@amazon.com}
\and
Zhi Zhang\\
Amazon Web Services \\
{\tt\small zhiz@amazon.com}
\and
Mu Li\\
Amazon Web Services \\
{\tt\small mli@amazon.com}
\and
Dimitris Metaxas\\
Rutgers University\\
{\tt\small dnm@cs.rutgers.edu}
}

\maketitle
\ificcvfinal\thispagestyle{empty}\fi

\begin{abstract}
    Traditional normalization techniques (e.g., Batch Normalization and Instance Normalization) generally and simplistically assume that training and test data follow the same distribution. As distribution shifts are inevitable in real-world applications, well-trained models with previous normalization methods can perform badly in new environments. Can we develop new normalization methods to improve generalization robustness under distribution shifts? In this paper, we answer the question by proposing CrossNorm and SelfNorm. CrossNorm exchanges channel-wise mean and variance between feature maps to enlarge training distribution, while SelfNorm uses attention to recalibrate the statistics to bridge gaps between training and test distributions. CrossNorm and SelfNorm can complement each other, though exploring different directions in statistics usage. Extensive experiments on different fields (vision and language), tasks (classification and segmentation), settings (supervised and semi-supervised), and distribution shift types (synthetic and natural) show the effectiveness. Code is available at \url{https://github.com/amazon-research/crossnorm-selfnorm}
\end{abstract}

\vspace{-8pt}
\section{Introduction}

Normalization methods, e.g., Batch Normalization \cite{ioffe2015batch}, Layer Normalization \cite{ba2016layer}, and Instance Normalization \cite{ulyanov2016instance}, play a pivotal role in training deep neural networks by making training more stable and convergence faster, assuming that training and test data come from the same distribution. However, distribution shifts in various real-world scenarios \cite{hendrycks2019benchmarking,richter2016playing,hendrycks2020pretrained} make traditional normalization techniques impractical. For instance, a driving scene segmentation model trained on one city usually does not generalize well to another city. In this paper, we aim to explore how normalization can improve generalization under distribution shifts. Specifically, we tackle the distribution shift problem from two respects: {\it enlarging training distribution} and {\it reducing test distribution}. 

\begin{figure*}[t]
\centering
\includegraphics[width=1\linewidth]{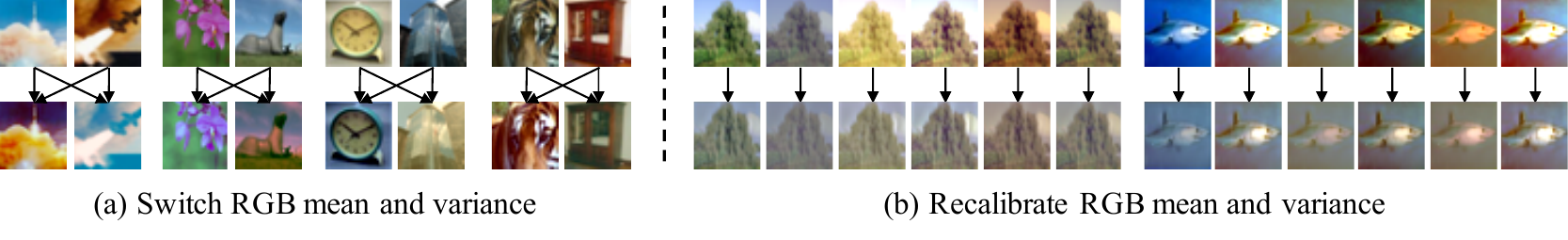}
\caption{Examples of exchanging ({\bf Left}) and adjusting ({\bf Right}) RGB mean and variance. Swapping the statistics can enrich image styles and thus enlarge training distribution, while recalibrating the statistics properly can encourage style consistency, reducing the train-test distribution gap.}
\label{fig:toy-examples}
\vspace{-10pt}
\end{figure*}

First, enlarging the training distribution is not in line with the conventional purpose of normalization which is to stabilize and accelerate training. So, can we employ normalization for a different goal--augmenting training data? Our inspiration comes from a simple observation that exchanging the RGB mean and variance between two images can transfer style between them, as shown in Figure \ref{fig:toy-examples} (a). For many tasks such as CIFAR image classification \cite{krizhevsky2009learning}, style, encoded by channel-wise mean and variance, is usually less critical in recognizing the object than other information, such as object shape. Therefore, augmenting style is safe enough that content labels remain unchanged. To augment style, we propose CrossNorm, which swaps channel-wise mean and variance of feature maps in training so that the model becomes more robust to changes in appearance.

Even with the augmented training data, a model will still encounter data with unforeseen appearances in deployment. Hence, another question comes: how to make normalization reduce test data distribution, i.e., bridging distribution gaps between training and test data? Similarly, our method is motivated by an observation illustrated in Figure \ref{fig:toy-examples} (b).
Given one image in different styles, we can reduce the style discrepancy when adjusting the RGB means and variances properly. Intuitively, style recalibration can reduce appearance variance so that training and test data will share more consistent styles. To this end, we propose SelfNorm by using attention \cite{hu2018squeeze} to adjust channel-wise mean and variance.

It is interesting to analyze the distinction and connection between CrossNorm and SelfNorm. At first glance, they take opposite actions (style augmentation vs. style reduction). Even so, they use the same tool: channel-wise statistics and pursue the same goal: generalization robustness. Additionally, CrossNorm can increase the capacity of SelfNorm by letting SelfNorm learn from more diverse styles in training. Overall, the key contributions are three-fold:

\begin{itemize}
\setlength\itemsep{0pt}
    \item Unlike previous efforts, we explore a new direction of using feature normalization for generalization under distribution shifts.
    \item We propose CrossNorm and SelfNorm, two simple yet effective normalization techniques that complement each other to improve generalization robustness.
    \item CrossNorm and SelfNorm can advance state-of-the-art robustness performance for different fields (vision or language), tasks (classification and segmentation), settings (fully or semi-supervised), and distribution shift types (synthetic and natural).
\end{itemize}

\vspace{-2ex}
\section{Related Work}


{\bf Generalization under synthetic distribution shifts}. Following the categorization in \cite{taori2020measuring}, distribution shifts are synthetic if they modify existing images to get shifted test datasets. Adversarial examples \cite{goodfellow2014explaining,madry2017towards} are one class of synthetic distribution shifts that were widely studied. Recently, various image corruptions \cite{hendrycks2019benchmarking}, as another synthetic type, have attracted increasing attention. 
To improve the robustness to corruptions, Stylized-ImageNet \cite{geirhos2019imagenet} conducts style augmentation to reduce the texture bias of CNNs. Recently, AugMix \cite{hendrycks2020augmix} trains robust models by mixing multiple augmented images based on random image primitives or image-to-image networks \cite{hendrycks2020many}. Adversarial noises training (ANT) \cite{rusak2020simple} and unsupervised domain adaptation \cite{schneider2020improving} can also improve the robustness against corruption. 

CrossNorm has two advantages over Stylized-ImageNet, though they are related. First, CrossNorm is efficient as it transfer styles directly in the feature space of target CNNs. However, Stylized-ImageNet relies on external style datasets and pre-trained style transfer models. Second, CrossNorm can advance the performance on both clean and corrupted data, while Stylized-ImageNet hurts clean generalization because external styles can result in massive training distribution shifts. Also,
CrossNorm is orthogonal to AugMix and ANT, making it possible for their joint usage. 





{\bf Generalization under natural distribution shifts}. Compared to synthetic distribution shifts, natural shifts refers to distribution gaps between unmodified data. One type of natural shifts is from video data, where adjacent frames are perceptually similar for humans, but they usually get inconsistent predictions from deep models \cite{gu2019using,shankar2019image}. Another type is dataset gaps \cite{recht2019imagenet,barbu2019objectnet,borji2020objectnet} arising from different factors, e.g., where and when, in collecting two separate datasets. For example, the semantic segmentation dataset GTA5 \cite{richter2016playing} comes from computer games, which naturally has distribution gaps with realistic segmentation datasets such as  Cityscape \cite{cordts2016cityscapes}. To address the distribution gaps, IBN \cite{pan2018two} mixes Instance and Batch Normalizations to narrow the distribution gaps. Domain randomization \cite{yue2019domain} uses style augmentation for domain generalization on segmentation datasets. It suffers from the same issues of Stylized-ImageNet as it also uses pre-trained style transfer models and additional style datasets. 

Compared to IBN and domain randomization, SelfNorm can bridge the distribution gaps with style recalibration, and CrossNorm is more efficient and balances better between the source and target datasets' performance. Beyond the vision field, natural language processing (NLP) applications also face the generalization challenges \cite{hendrycks2020pretrained} posed by distribution shifts. Fortunately, SelfNorm and CrossNorm can also improve model robustness in NLP.



{\bf Normalization and attention}. 
Batch Normalization \cite{ioffe2015batch} is a milestone technique that inspires many following normalization methods such as Instance Normalization \cite{ulyanov2016instance}, Layer Normalization \cite{ba2016layer}, and Group Normalization \cite{wu2018group}. Recently, some works integrate attention \cite{hu2018squeeze} into normalization. Mode normalization \cite{deecke2018mode} and attentive normalization \cite{li2019attentive} use attention to weigh a mixture of Batch Normalizations. IEBN \cite{liang2020instance} uses attention to regulate the batch noises in Batch Normalization. Examplar Normalization \cite{zhang2020exemplar} learns to combine multi-type normalizations by attention. By contrast, SelfNorm uses attention with only Instance Normalization. With attention's help, SelfNorm can emphasize important styles and suppress trivial ones, reducing the distribution gaps caused by appearance discrepancy.


\begin{figure*}[t]
\centering
\includegraphics[width=.9\linewidth]{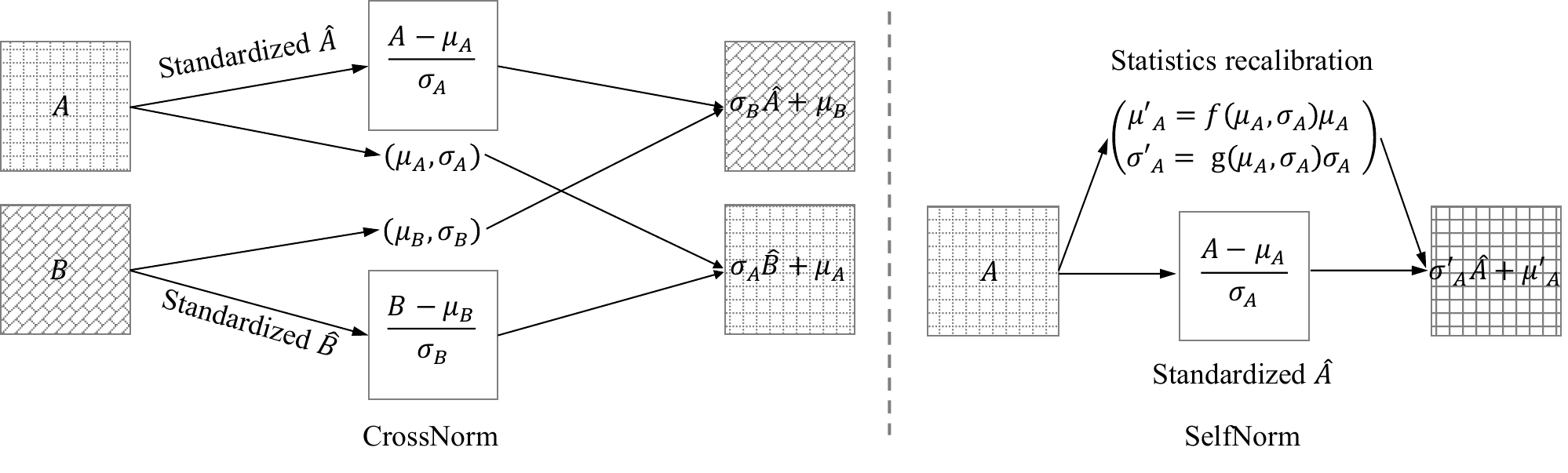}
\caption{CrossNorm ({\bf left}) and SelfNorm ({\bf right}). CrossNorm swaps the mean and variance between a pair of feature maps, while SelfNorm uses attention to recalibrate a feature map's statistics.}
\label{fig:framework}
\vspace{-10pt}
\end{figure*}

{\bf Data augmentation}. Data augmentation is an important tool in training deep models. Current popular data augmentation techniques are either label-preserving \cite{cubuk2019autoaugment, lim2019fast,ho2019population} or label-perturbing \cite{zhang2017mixup,yun2019cutmix}. The label-preserving methods usually rely on domain-specific image primitives, e.g., rotation and color, making them inflexible for tasks beyond the vision field. The label-perturbing techniques mainly work for classification and may have trouble in broader applications, e.g., segmentation. CrossNorm, as a data augmentation method, is readily applicable to diverse fields (vision and language) and tasks (classification and segmentation). The goal of CrossNorm is to boost generalization robustness under distribution shifts, which is also different from many former data augmentation methods.




\vspace{-5pt}
\section{CrossNorm and SelfNorm}

This section elaborates CrossNorm, SelfNorm, their relation, and their application in deep neural networks. Before that, we introduce some preliminaries regarding Instance Normalization \cite{ulyanov2016instance} and the style concept.

\subsection{Preliminary} 

{\bf Instance Normalization}. Technically, SelfNorm and CrossNorm share the same origin: Instance Normalization \cite{ulyanov2016instance}.  In 2D CNNs, each instance has $C$ feature maps of size $H\times W$. Given a feature map $\mathcal{A}\in \R^{H\times W}$, Instance Normalization first normalizes the feature map and then conducts affine transformation:
\begin{equation}\label{eq:instancenorm}
    \gamma\frac{\mathcal{A}-\mu_\mathcal{A}}{\sigma_\mathcal{A}}+\beta,
\end{equation}
where $\mu_\mathcal{A}$ and $\sigma_\mathcal{A}$ are the mean and standard deviation; $\gamma$ and $\beta$ denotes learnable affine parameters. As shown in Figure \ref{fig:toy-examples} and also pointed out by the style transfer practices \cite{dumoulin2016learned,ulyanov2017improved,huang2017arbitrary}, $\mu_\mathcal{A}$ and $\sigma_\mathcal{A}$ can encode some style information.

{\bf Style Concept}. In this paper, the style concept refers to a family of weak cues associated with the semantic content of interest. For instance, the image style in object recognition can include many appearance-related factors such as color, contrast, and brightness. Style sometimes may help in decision-making, but the model should rely more on vital content cues to become robust. To reduce its bias rather than discarding it, we use CrossNorm with probability in training. The insight beneath CrossNorm is that each instance, or feature map, has its unique style. Further, style cues are not equally important. For example, the yellow color seems more useful than other style cues in recognizing an orange. In light of this, the intuition behind SelfNorm is that attention may help emphasize essential styles and suppress trivial ones. Although we use the channel-wise mean and variance to modify styles, we do not assume that they are sufficient to represent all style cues. Better style representations are available with more complex statistics \cite{li2017universal} or even style transfer models \cite{ulyanov2017improved, huang2017arbitrary}. We choose the first and second-order statistics mainly because they are simple, efficient to compute, and can naturally connect normalization to generalization robustness.

\subsection{CrossNorm} 
To enlarge training distribution, CrossNorm exchanges $\mu_\mathcal{A}$ and $\sigma_\mathcal{A}$ of channel $A$ with $\mu_\mathcal{B}$ and $\sigma_\mathcal{B}$ of channel $\mathcal{B}$, i.e., changing $\beta$ and $\gamma$ to each other's $\mu$ and $\sigma$, seen in Figure \ref{fig:framework}:
\begin{equation}\label{eq:crossnorm}
    \sigma_\mathcal{B}\frac{\mathcal{A}-\mu_\mathcal{A}}{\sigma_\mathcal{A}}+\mu_\mathcal{B}\qquad	\sigma_\mathcal{A}\frac{\mathcal{B}-\mu_\mathcal{B}}{\sigma_\mathcal{B}}+\mu_\mathcal{A},
\end{equation}
where {\it $\mathcal{A}$ and $\mathcal{B}$ seem to normalize each other, hence CrossNorm.} CrossNorm is motivated by the key observation that a target dataset, such as a classification dataset, has rich, though subtle, styles. Specifically, each instance, or even every channel, has its unique style. CrossNorm, turned on with some probability in training, can perform efficient style augmentation and thus enlarge training distribution. To further diversify styles, we investigate different feature map choices, resulting in different CrossNorm variants.


\begin{figure*}[t]
\centering
\includegraphics[width=0.8\linewidth]{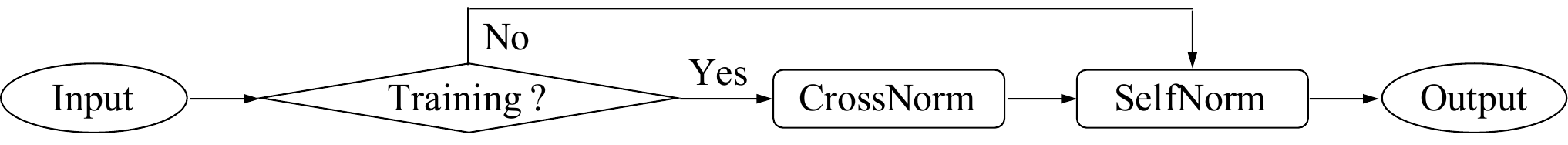}
\caption{Flowchart for CrossNorm and SelfNorm. CrossNorm works only in training, while SelfNorm learns in training and functions in testing.}
\label{fig:sn-cn-flowchart}
\vspace{-9pt}
\end{figure*}

 
{\bf 1-instance mode}. For 2D CNNs, given one instance $\mathcal{X} \in R^{C\times H\times W}$, CrossNorm can exchange statistics between its $C$ channels:
\begin{equation}
	\left\{(\mathcal{A},\mathcal{B})\in(\mathcal{X}_{i, :, :},\mathcal{X}_{j, :, :})\mid i\neq j, 0<i,j<C \right\},
\end{equation}
where $\mathcal{A}$ and $\mathcal{B}$ refer to the channel pair in Equation \ref{eq:crossnorm}. 

{\bf 2-instance mode}. If two instances $\mathcal{X}, \mathcal{Y}\in R^{C\times H\times W}$ given, CrossNorm can swap statistics between their corresponding channels, i.e., $\mathcal{A}$ and $\mathcal{B}$ become:
\begin{equation}
	\left\{(\mathcal{A},\mathcal{B})\in(\mathcal{X}_{i, :, :},\mathcal{Y}_{i, :, :})\mid 0<i<C \right\}.
\end{equation}
Compared to 1-instance CrossNorm, 2-instance CrossNorm considers instance-level style instead of channel-level.

{\bf Crop}. Moreover, distinct spatial regions probably have different mean and variance statistics. To promote the style diversity, we propose to crop regions for CrossNorm: 
\begin{equation}
	\left\{(\mathcal{A},\mathcal{B})\in(\text{crop}(\mathcal{A}),\text{crop}(\mathcal{B}))\mid r_{\text{crop}} \geq t\right\}
\end{equation}
where the crop function returns a square with area ratio $r$ no less than a threshold $t(0<t\leq1)$. The whole channel is a special case in cropping. There are three cropping choices: content only, style
only, and both. For content cropping, we crop A only when we use its standardized feature map. In other words, no cropping applies to A when it provides its statistics to B. Cropping both means cropping A
and B no matter we employ their standardized feature map or statistics. The cropping strategy can produce diverse
styles for both the 1-instance and 2-instance CrossNorms.

\subsection{SelfNorm}
To bridge the train-test distribution gap, SelfNorm replaces $\beta$ and $\gamma$ in Equation \ref{eq:instancenorm} with recalibrated mean $\mu'_\mathcal{A}=f(\mu_\mathcal{A}, \sigma_\mathcal{A})\mu_\mathcal{A}$ and standard deviation $\sigma'_\mathcal{A}=g(\mu_\mathcal{A}, \sigma_\mathcal{A})\sigma_\mathcal{A}$, as illustrated in Figure \ref{fig:framework}, where $f$ and $g$ are the attention functions. The adjusted feature map becomes:
\begin{equation}\label{eq:selfnorm}
    \sigma'_\mathcal{A}\frac{\mathcal{A}-\mu_\mathcal{A}}{\sigma_\mathcal{A}}+\mu'_\mathcal{A}.
\end{equation}
As $f$ and $g$ learn to scale $\mu_\mathcal{A}$ and $\sigma_\mathcal{A}$ based on themselves, {\it $\mathcal{A}$ normalizes itself by self-gating, hence SelfNorm}. SelfNorm is inspired by the fact that attention can help the model emphasize informative features and suppress less useful ones. In terms of recalibrating $\mu_\mathcal{A}$ and $\sigma_\mathcal{A}$, SelfNorm expects to highlight the discriminative styles shared by training and test distributions and understate trivial one-sided styles. In practice, we use two fully connected (FC) networks to wrap attention functions $f$ and $g$, respectively. Each network is  efficient as the input and output are two and one scalars.


Note that SelfNorm is different from SE \cite{hu2018squeeze}, though they use similar attention. First, SE models the interdependency between channels, while SelfNorm deals with each channel independently. Second, SelfNorm learns to recalibrate channel-wise mean and variance instead of channel features in SE. Also, a SelfNorm unit, with complexity $O(C)$, is more lightweight than a SE one, of $O(C^2)$, where $C$ denotes the channel number. 


\subsection{Relation and Application}
{\bf Unity of opposites}.  CrossNorm and SelfNorm both start from Instance Normalization but head in opposite directions. CrossNorm transfers statistics between channels, enriching the combinations of standardized features (zero-mean and unit-variance) and statistics. In contrast, SelfNorm recalibrates statistics to focus on only necessary styles, reducing standardized features and statistics mixtures' diversity. They perform opposite operations mainly because they target different stages. CrossNorm functions only in training, whereas SelfNorm dedicates to style recalibration during testing. Note that SelfNorm is a learnable module, requiring training to work. Figure \ref{fig:sn-cn-flowchart} shows the flowchart of CrossNorm and SelfNorm. Despite these differences, they both can facilitate generalization under distribution shifts. Further, CrossNorm can boost SelfNorm's performance because its style augmentation can prevent SelfNorm from overfitting to specific styles. Overall, the two seemingly opposed methods form a unity of using normalization statistics to advance generalization robustness.


{\bf Modular design}. CrossNorm and SelfNorm can naturally work in the feature space, making it flexible to plug them into many network locations. Two questions arise: how many units are necessary and where to place them? To simplify the questions, we turn to the modular design by embedding them into a network cell. For example, in ResNet \cite{he2016deep}, we put them into a residual module. The search space significantly shrinks for the limited positions in a residual module. We will investigate the position choices in experimental ablation study. The modular design allows using multiple CrossNorms and SelfNorms in a network. We will show in the ablation study that accumulated style recalibrations are helpful for model robustness. 

\section{Experiment}
We evaluate CrossNorm (CN) and SelfNorm (SN) in various distribution shifts settings.

\begin{table*}[t]
\begin{center}
\caption{mCE (\%) on CIFAR-10-C and CIFAR-100-C. CNSN obtains lower errors than most previous methods with different backbones. Albeit some higher errors than AugMix, it is more general without relying on the image primitives, e.g., rotation, in AugMix. As CNSN and AugMix are orthogonal, their joint usage brings new state-of-the-art results.}\label{tb:robust-cifar}\vspace{-1ex}
\setlength\tabcolsep{3pt}
\begin{tabular}{l|ccccccc|cccc}
\toprule
CIFAR-10-C & Basic & Cutout & Mixup & CutMix & AutoAug & AdvTr. & AugMix & CN & SN & CNSN & CNSN+AugMix\\
\hline
AllConvNet & 30.8 & 32.9 & 24.6 & 31.3 & 29.2 & 28.1 & 15.0 & 26.0 & 24.0 & 17.2 & {\bf 11.8}\\
DenseNet & 30.7 & 32.1 & 24.6 & 33.5 & 26.6 & 27.6 & 12.7 & 24.7 & 22.0 & 18.5 & {\bf 10.4}\\
WideResNet & 26.9 & 26.8 & 22.3 & 27.1 & 23.9 & 26.2 & 11.2 & 21.6 & 20.8 & 16.9 & {\bf 9.9}\\
ResNeXt & 27.5 & 28.9 & 22.6 & 29.5 & 24.2 & 27.0 & 10.9 & 22.4 & 21.5 & 15.7 & {\bf 9.1}\\
\hline
Mean & 29.0 & 30.2 & 23.5 & 30.3 & 26.0 & 27.2 & 12.5 & 23.7 & 22.1 & 17.0 & {\bf 10.3}\\
\toprule
CIFAR-100-C & Basic & Cutout & Mixup & CutMix & AutoAug & AdvTr. & AugMix & CN & SN & CNSN & CNSN+AugMix\\
\hline
AllConvNet & 56.4 & 56.8 & 53.4 & 56.0 & 55.1 & 56.0 & 42.7 & 52.2 & 50.3 & 42.8 & {\bf 36.8}\\
DenseNet & 59.3 & 59.6 & 55.4 & 59.2 & 53.9 & 55.2 & 39.6 & 55.4 & 53.9 & 48.5 & {\bf 37.0}\\
WideResNet & 53.3 & 53.5 & 50.4 & 52.9 & 49.6 & 55.1 & 35.9 & 48.8 & 47.4 & 43.7 & {\bf 33.4}\\
ResNeXt & 53.4 & 54.6 & 51.4 & 54.1 & 51.3 & 54.4 & 34.9 & 47.0 & 47.6 & 40.8 & {\bf 30.8}\\
\hline
Mean & 55.6 & 56.1 & 52.6 & 55.5 & 52.5 & 55.2 & 38.3 & 50.9 & 49.8 & 43.5 & {\bf 34.7}\\
\bottomrule
\end{tabular}
\end{center}
\vspace{-1em}
\end{table*}

\begin{table*}[t]
\begin{center}
\caption{Clean error and mCE (\%) of ResNet50 trained 90 epochs on ImageNet. CNSN, using simple general statistics, achieves comparable performance as domain-specific AugMix. Jointly applying CNSN with AugMix and IBN can produce the lowest clean and corruption errors.}\label{tb:robust-imagenet-c}\vspace{-1ex}
\small
\setlength\tabcolsep{1.5pt}
\begin{tabular}{l|c|ccc|cccc|cccc|cccc|c}
\toprule
& & \multicolumn{3}{c}{Noise} & \multicolumn{4}{c}{Blur} & \multicolumn{4}{c}{Weather} & \multicolumn{4}{c}{Digital} \\
\hline
Aug. & Clean & Gauss. & Shot & Impulse & Defocus & Glass & Motion & Zoom & Snow & Frost & Fog & Bright & Contrast & Elastic & Pixel & JPEG & mCE\\
\hline
Standard & 23.9 & 79 & 80 & 82 & 82 & 90 & 84 & 80 & 86 & 81 & 75 & 65 & 79 & 91 & 77 & 80 & 80.6\\
Patch Uniform & 24.5 & 67 & 68 & 70 & 74 & 83 & 81 & 77 & 80 & 74 & 75 & 62 & 77 & 84 & 71 & 71 & 74.3\\
Random AA* & 23.6 & 70 & 71 & 72 & 80 & 86 & 82 & 81 & 81 & 77 & 72 & 61 & 75 & 88 & 73 & 72 & 76.1\\
MaxBlur pool & 23.0 & 73 & 74 & 76 & 74 & 86 & 78 & 77 & 77 & 72 & 63 & 56 & 68 & 86 & 71 & 71 & 73.4\\
SIN & 27.2 & 69 & 70 & 70 & 77 & 84 & 76 & 82 & 74 & 75 & 69 & 65 & 69 & 80 & 64 & 77 & 73.3\\
AugMix* & 23.4 & 66 & 66 & 66 & 69 & 80 & 65 & 68 & 72 & 72 & 66 & 60 & 63 & 78 & 66 & 71 & 68.4\\
\hline
CN & 23.3 & 73 & 75 & 75 & 78 & 89 & 79 & 82 & 79 & 75 & 66 & 61 & 69 & 97 & 69 & 74 & 75.1 \\
SN & 23.7 & 69 & 71 & 69 & 77 & 87 & 77 & 80 & 75 & 77 & 70 & 61 & 73 & 83 & 61 & 71 & 73.8\\
CNSN & 23.3 & 66 & 67 & 65 & 77 & 89 & 76 & 80 & 72 & 72 & 67 & 59 & 47 & 83 & 62 & 72 & 69.7 \\
CNSN+AugMix & {\bf 22.3} & 61 & 62 & 60 & 70 & 77 & 62 & 68 & 62 & 65 & 63 & 55 & 43 & 73 & 55 & 66 & {\bf 62.8}\\
\bottomrule
\end{tabular}
\end{center}
\vspace{-2em}
\end{table*}

{\bf Image classification datasets.} We use benchmark datasets: CIFAR-10 \cite{krizhevsky2009learning}, CIFAR-100, and ImageNet\cite{deng2009imagenet}. To evaluate the model robustness against corruption, we use the datasets: CIFAR-10-C, CIFAR-100-C, and ImageNet-C \cite{hendrycks2019benchmarking}. These datasets are the original test data poisoned by 15 everyday image corruptions from 4 general types: noise, blur, weather, and digital. Each noise has 5 intensity levels when injected into images. In addition, we conduct domain adaptation experiments with Office-31 including 4652 images and 31 categories from 3 domains: Amazon (A) ( amazon.com images), Webcam (W) ( web camera images) and DSLR (D) (digital SLR camera images). 

{\bf Image segmentation datasets.} We further validate our method using a domain generalization setting, where the models are trained in a source domain and tested on a unforseen target domain. We use the synthetic dataset Grand Theft Auto V (GTA5) \cite{richter2016playing} as the source domain and generalize to the real-world dataset Cityscapes \cite{cordts2016cityscapes}. GTA5 has the training, validation, and test divisions of 12,403, 6,382, and 6,181, more than those of 2,975, 500, and 1,525 from Cityscapes. Despite the differences, their pixel categories are compatible with each other, allowing to evaluate models' generalization capability from one to another.
{\bf Sentiment classification datasets.} Besides vision tasks, we demonstrate that our method can work well on NLP tasks. In particular, we use the cross-dataset binary sentiment classification setting, where a model is trained on the IMDb dataset \cite{maas2011learning} and then tested on the SST-2 dataset \cite{socher2013recursive}. The IMDb dataset collects highly polarized full-length lay movie reviews with 25,000 positive and 25,000 negative reviews. The SST-2, with 9613 and 1821 reviews for training and testing, is also a binary sentiment dataset but instead contains pithy expert movie reviews.  





{\bf Metric.} For image classification, we use test errors to measure robustness. Given corruption type $c$ and severity $s$, let $E^c_s$ denote the test error. For CIFAR datasets, we use the average over 15 corruptions and 5 severities: $1/75\sum_{c=1}^{15}\sum_{s=1}^{5}E_{c,s}$. In contrast, for ImageNet, we normalize the corruption errors by those of AlexNet \cite{krizhevsky2012imagenet}: $1/15\sum_{c=1}^{15}(\sum_{s=1}^{5}E^c_s/\sum_{s=1}^{5}E_{c,s}^{AlexNet})$. The above two metrics follow the convention \cite{hendrycks2020augmix} and are denoted as mean corruption errors (mCE) whether they are normalized or not. Different from classification, segmentation uses the mean Intersection over Union (mIoU) of all categories. For sentiment classification, we report accuracy as its metric.

{\bf Hyper-parameters.} In the experiments, an attention function in SN uses one fully connected layer, followed by Batch Norm and a sigmoid layer. We put CN ahead of SN, and plug them into every cell in a network, e.g., each residual module in a ResNet. During training, we turn on only some CNs with probability to avoid excessive data augmentation. Unless specified, 2-instance CN is used with cropping. We sample the cropping bounding box ratio uniformly and set the threshold $t=0.1$. Refer to the appendix for details regarding CN's active number and probability.

\begin{table}[t]
\begin{center}
\caption{
Semi-supervised results on CIFAR-10 with WideResNet-28-2. We use FixMatch with weak augmentation (WA) or strong RandAugment (RA). In either case, CN substantially reduces both clean and corruption errors.
}\label{tb:semi-supervised}\vspace{-1ex}
\small
\setlength\tabcolsep{4pt}
\begin{tabular}{l|l|cc|cc}
\toprule
& & \multicolumn{2}{c|}{FixMatch+WA} & \multicolumn{2}{c}{FixMatch+RA} \\
\hline
& & Baseline & +CN & Baseline & +CN \\ 
\hline
250 & Clean err(\%) & 57.7 & {\bf 51.4} & 10.9 & {\bf 7.4}\\
 labels & mCE(\%) & 62.9 & {\bf 56.5} & 23.4 & {\bf 16.8}\\
\hline
1000  & Clean err(\%) & 16.7 & {\bf 13.1} & 6.7 & {\bf 5.8}\\
labels & mCE(\%) & 37.7 & {\bf 29.6} & 19.8 & {\bf 15.5}\\
\bottomrule
\end{tabular}
\end{center}
\vspace{-2.5em}
\end{table}





\vspace{-0.5ex}
\subsection{Robustness against Unseen Corruptions}
{\bf Supervised training on CIFAR.} Following AugMix \cite{hendrycks2020augmix}, we use four different backbones: an All Convolutional Network \cite{springenberg2014striving}, a DenseNet-BC (k = 12, d = 100) \cite{huang2017densely}, a 40-2 Wide ResNet \cite{zagoruyko2016wide}, and a ResNeXt-29 (32$\times$4) \cite{xie2017aggregated}. CN and SN are plugged into their cell modules. We use the same hyper-parameters in the AugMix Github repository\footnote{\label{augmix-repo}https://github.com/google-research/augmix}. 


According to Table \ref{tb:robust-cifar}, individual CN and SN can outperform most previous approaches on robustness against unseen corruptions and combining them can decrease the mean error by $\sim$12\% on both CIFAR-10-C and CIFAR-100-C. As the corruptions mainly change image textures, one possible explanation is that CN and SN, through style augmentation and recalibration, may help reduce the texture sensitivity and bias, making the classifiers more robust to unseen corruptions. Also, CN and SN are orthogonal to AugMix, which relies on domain-specific image operations. Their joint application can continue to lower the mCEs by 2.2\% and 3.6\% on top of AugMix.



{\bf Supervised training on ImageNet.} Following the AugMix Github repository, we train a ResNet-50 for 90 epochs with weight decay 1e-4. The learning rate starts from 0.1, divided by 10 at epochs 30 and 60. Note that AugMix reports the results of 180 epochs in their paper. For a fair comparison, we also train it 90 epochs in our experiments. Different from CIFAR experiments, we apply CN only to the image space. Besides, we also add Instance-batch normalization (IBN) \cite{pan2018two} in the final combination with AugMix. It was initially designed for domain generalization but can also boost model robustness against corruption.


Table \ref{tb:robust-imagenet-c} gives the results on ImageNet. We can observe that both clean and corrupted errors decrease when applying CN and SN separately. Their joint usage can make the clean and corruption errors drop by 0.6\% and 10.9\% simultaneously, closing the gap with AugMix. Moreover, applying CN and SN on top of AugMix can significantly lower its clean and corruption errors by 1.1\% and 5.6\%, respectively, achieving state-of-the-art performance. IBN also makes some contributions here since it is complementary to other components.

\begin{table}[t]
\begin{center}
\caption{Image segmentation results (mIoU) on GTA5-Cityscapes domain generalization using a FCN with ResNet50. CN and SN are comparable to domain randomization (DR) and IBN on the target domain (Cityscapes). Combining CN and SN can achieve state-of-the-art results.}\label{tb:segmentation}\vspace{-1ex}
\setlength\tabcolsep{4pt}
\begin{tabular}{l|ccc|ccccc}
\toprule
Methods & Baseline & IBN & DR & CN & SN & CNSN \\
\midrule
Source  & 63.7 & 64.2 & 49.0  & 61.2 &\textbf{64.6}  &63.5\\
\hline
Target & 21.4 & 29.6 & 32.7 & 32.0  &29.9 &\textbf{36.5}\\
\bottomrule
\end{tabular}
\end{center}
\vspace{-2.5em}
\end{table}
{\bf Semi-supervised training on CIFAR.} Apart from supervised training, we also evaluate CN in semi-supervised learning.  Following state-of-the-art FixMatch \cite{sohn2020fixmatch} setting, we train a 28-2 Wide ResNet for 1024 epochs on CIFAR-10. The SGD optimizer applies with Nesterov momentum 0.9, learning rate 0.03, and weight decay 5e-4. The probability threshold to generate pseudo-labels is 0.95, and the weight for unlabeled data loss is 1. We sample 250 and 4,000 labeled data with random seed 1, leaving the rest as unlabeled data. In each experiment, we apply CrossNorm to either all data or only unlabeled data and choose the better one. Our experiments use the Pytorch FixMatch implementation \footnote{https://github.com/kekmodel/FixMatch-pytorch}.



Table \ref{tb:semi-supervised} shows the semi-supervised results. Whether FixMatch uses only weak random flip and crop augmentations or strong RandAugment \cite{cubuk2019randaugment}, CN can always decrease both the clean and corruption errors, demonstrating its effectiveness in semi-supervised training. Especially with the help of CN, training with 250 labels even has 3\% lower corruption error than with 1000 labels, according to the columns 5 and 6. Additionally, two points are noteworthy here. First, we try FixMatch with only weak augmentations to simulate more general situations. For new areas other than natural images, humans may have the limited expertise to design advanced augmentation operations. Fortunately, CN is area-agnostic and easily applicable to such situations.
Moreover, previous semi-supervised methods mainly focus on clean generalization. Here we introduce corruption robustness as another metric for comprehensive evaluation. 

\vspace{-0.5ex}
\subsection{Generalization from Synthetic to Realistic Data}



\textbf{Setup.} We perform cross-dataset generalization from GTA5 (synthetic) to Cityscapes (realistic), following the setting of IBN \cite{pan2018two}. It uses 1/4 training data in GTA5 to match the data scale of Cityscapes. We train the FCN \cite{long2015fully} with ResNet50 backbone in source domain GTA5 for 80 epochs with batch size 16. The network is initialized with ImageNet pre-trained weights. We test the trained model on both the source and target domains. The training uses random scaling, flip, rotation, and cropping ($713\times713$) for data augmentation. We use the 2-instance CN with style cropping in this setting. Besides, we re-implement the domain randomization \cite{yue2019domain} and make the training iterations the same as ours. It transfers the synthetic images to 15 auxiliary domains with ImageNet image styles.


\textbf{Results.} Table \ref{tb:segmentation} shows that CN and SN both can substantially increase the segmentation accuracy on the target domain by 10.6\% and 8.5\%. CN performs style augmentation to make the model focus more on domain-invariant features. SN learns to highlight the discriminative styles that are likely to share across domains. CN and SN get comparable generalization performance as state-of-the-art domain randomization \cite{yue2019domain} and IBN \cite{pan2018two}. However, CN significantly outperforms the domain randomization method by 12.2\% on the source accuracy because the domain randomization transfers external styles to the source training data, causing dramatic distribution shifts. Moreover, combining CN and SN gives the best generalization performance while still maintaining high source accuracy.

\begin{table}[t]
\begin{center}
\caption{Sentiment classification accuracy with cross-dataset generalization (IMDb\textrightarrow SST-2) using GloVe embedding and ConvNets. CN and SN work in the NLP field. }\label{tb:nlp}\vspace{-1ex}
\begin{tabular}{l|c|ccccc}
\toprule
Methods & Baseline & CN & SN & CNSN \\
\midrule
Source  & 85.7 & 85.1 &\textbf{86.3}  &85.9\\
\hline
Target & 71.9  & 73.0  &73.9  &\textbf{74.9}\\
\bottomrule
\end{tabular}
\end{center}
\vspace{-2.5em}
\end{table}

\vspace{-0.5ex}
\subsection{Cross-dataset Generalization in NLP}


\textbf{Setup.} To show that CN and SN are independent of application fields, we also evaluate their generalization robustness on a binary sentiment classification setup in the NLP area. The model is trained on the IMDb dataset and tested on the SST-2 dataset.  Follow the setting of \cite{hendrycks2020pretrained}, we use the GloVe \cite{pennington2014glove} word embedding and Convolutional Neural Networks (ConvNets) \cite{kim2014convolutional} as the classification model. We use the implementation of ConvNets in this repository\footnote{https://github.com/bentrevett/pytorch-sentiment-analysis}. The convolutional layers with three kernel sizes (3,4,5) are used to extract $n-gram$ features within the review texts.  CN and SN units are placed between the embedding layer and the convolutional layers. We use the Adam optimizer and train the model for 20 epochs.

\textbf{Results.} From Table \ref{tb:nlp}, we can find that SN improves the performance in both the source and target domains by 0.6\% and 2.0\%. CN can also increase target accuracy without much degradation in the source domain. Combining them gives a 3.0\% boost of target accuracy. This experiment indicates that CN and SN can also work in the NLP area, not limited to the vision tasks. Despite the lack of intuitive explanations as for the image data, the mean and variance statistics in NLP data are also useful in facilitating generalization under distribution shifts.



\vspace{-0.5ex}
\subsection{Domain Adaptation}

\begin{table}[t]
\begin{center}
\caption{Domain adaptation accuracy on Office-31. CN outperforms AdaBN in 4 settings and mean accuracy. }\label{tb:domain-adaptation}\vspace{-1ex}
\small
\setlength\tabcolsep{1pt}
\begin{tabular}{l|ccccccc}
\toprule
& A\textrightarrow W & D\textrightarrow W & W\textrightarrow D & A\textrightarrow D & D\textrightarrow A & W\textrightarrow A & Avg.\\
\midrule
Source Only & 68.4 & 96.7 & 99.3 & 68.9 & 62.5 & 60.7 & 76.1\\
AdaBN & 74.1 & 97.1 & 99.5 & 72.3 & {\bf 61.9} & {\bf 61.2} & 77.7\\
CN & {\bf 77.6} & {\bf 98.0} & {\bf 100.0} & {\bf 77.9} & 60.9 & 61.0 & {\bf 79.2}\\
\bottomrule
\end{tabular}
\end{center}
\vspace{-2em}
\end{table}

{\bf Setup.} In addition to generalization, we evaluate CN on domain adaptation. In particular, we compare CN with closely related AdaBN \cite{li2016revisiting} in 6 adaptation settings of the Office-31 dataset. CN is applied to both image and feature space without cropping. We follow a Github repo\footnote{https://github.com/fazilaltinel/ADDA.PyTorch-resnet} to use ResNet50, 100 epochs, batch size 32, and Adam optimizer with constant learning rate 1e-5 and weight decay 2.5e-5. 

{\bf Results.} According to Table \ref{tb:domain-adaptation}, CN improves 3.1\% average accuracy over the baseline, which almost doubles the improvements (1.6\%) of AdaBN. We notice that CN and AdaBN make accuracy increase slightly or even decrease in the D\textrightarrow A and  W\textrightarrow A settings. This may be due to that D (498 labeled images) and W (795 labeled images) domains have much fewer images than 
A domain (2817 images). 

\vspace{-0.5ex}
\subsection{Visualization}

Apart from the quantitative comparisons, we also provide some visualization results of CN and SN to better understand their effects. To this end, we map the feature changes made by CN and SN back to image space by inverting the feature representations \cite{mahendran2015understanding}. For detailed experimental settings, refer to the appendix.


\begin{figure*}[t]
\centering
\includegraphics[width=0.97\linewidth]{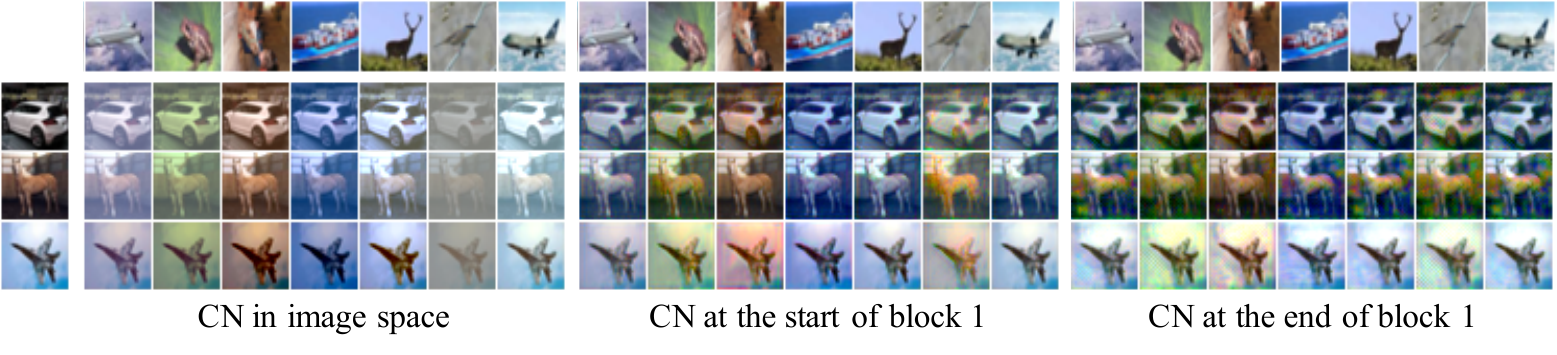}
\vspace*{-3mm}
\caption{CN visualization at image level ({\bf Left}), the head ({\bf Middle}) and tail ({\bf Right}) of block 1 in WideResNet-40-2. Both the content ({\bf Row}) and style ({\bf Column}) images are from CIFAR-10. The style rendering changes from global to local as CN gets deeper in the network.}
\label{fig:cn-vis}
\vspace{-1em}
\end{figure*}

\begin{figure*}[t]
\centering
\includegraphics[width=0.97\linewidth]{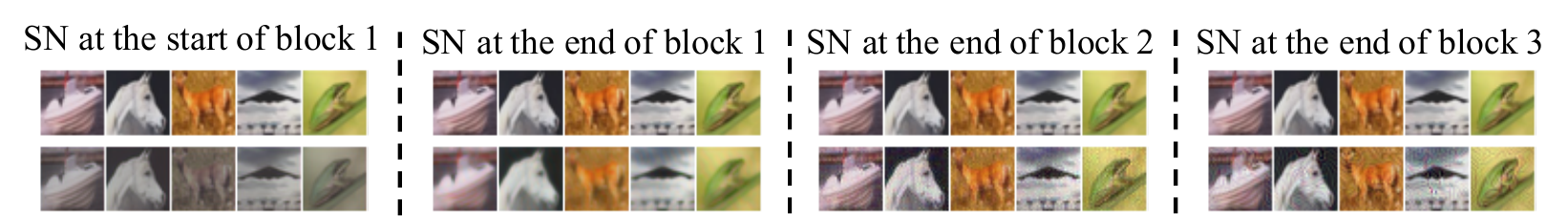}
\vspace*{-2mm}
\caption{Visualizing 4 single SNs in WideResNet-40-2 by comparing images before ({\bf Top}) and after ({\bf Bottom}) SN. The left two, lying in shallow locations, can adjust styles by suppressing color and adding blur. As SN goes deeper, the recalibration effect becomes subtle because the statistics of high-level features do not directly connect to low-level visual cues.}
\label{fig:single-sn-vis}
\vspace{-1em}
\end{figure*}


\begin{figure*}[t!]
\centering
\includegraphics[width=0.97\linewidth]{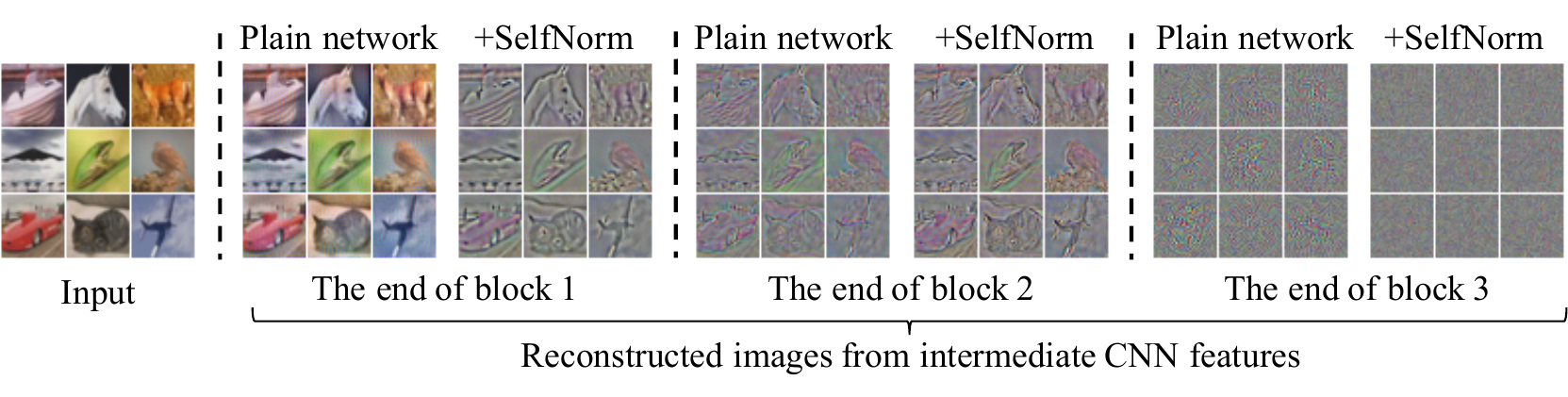}
\vspace*{-3mm}
\caption{Visualizing accumulated SNs in WideResNet-40-2 by comparing reconstructed images from intermediate features. SNs in block 1 can wash away much style information preserved in the vanilla network. Similarly, the plain network's final representation retains some high-frequency signals which are suppressed by SNs.}
\label{fig:combined-sn-vis}
\vspace{-1.5em}
\end{figure*}

In visualizing CN, we pair one content image with multiple style images for better illustration. We first forward them to get their feature representations at a chosen position. Then we compute standardized features from the content image representation and means and variances of the style image representations. The optimization starts from the content image and tries to fit its representation to the target one mixing the standardized features with different means and variances. Figure \ref{fig:cn-vis} shows diverse style changes made by CN. The style changes become more local and subtle as CN moves deeper in the network.

To visualize SN at a network location, we first forward an image to obtain the target representation immediately after the SN. Then we turn off the chosen SN and optimize the original image to make its representation fit the target one. In this way, we can examine a SN's effect by observing the changes in image space. As shown in Figure \ref{fig:single-sn-vis}, SN can primarily reduce the contrast and color at the first network block. The effect becomes more subtle as SN goes deeper into the network. One possible explanation is that the high-level representations lose too many low-level details, making it difficult to visualize the changes.

In addition to visualizing individual SNs, it is also interesting to see their compound effect. To this end, we reconstruct an image from random noises by matching its representation with a given one. The reconstructed image can show what information is preserved by the feature representation. By comparing two reconstructed images from a network with or without SN, we can observe the joined recalibration effects of SNs before a selected location. From Figure \ref{fig:combined-sn-vis}, we can find SNs in the first two network blocks can suppress much style information and preserve object shapes. The reconstructions from block 3 do not look visually informative due to the high-level abstraction. Even so, SNs can restrain the high-frequency signals kept in the vanilla network.

\vspace{-3mm}
\section{Conclusion}
This paper has explored how normalization can enhance generalization under distribution shifts and presented CN and SN, two simple, effective, and complementary normalization techniques. Their extensive applications can shed light on developing general methods applicable to multiple fields, such as vision and language, and broad synthetic and natural distribution shift circumstances. Given the simplicity of CN and SN, we believe there is substantial room for improvement. One possible direction is to explore better style representations since the current channel-wise mean and variance are not optimal to encode diverse styles.



{\small
\bibliographystyle{ieee_fullname}
\bibliography{main}
}
\clearpage
\vspace{-5em}
\section{Appendix}
In each experiment, we conduct a grid search on four CN configurations: active numbers (1, 2) and probability (0.25, 0.5), and report the best result.


\begin{figure*}[t]
\centering
\includegraphics[width=.9\linewidth]{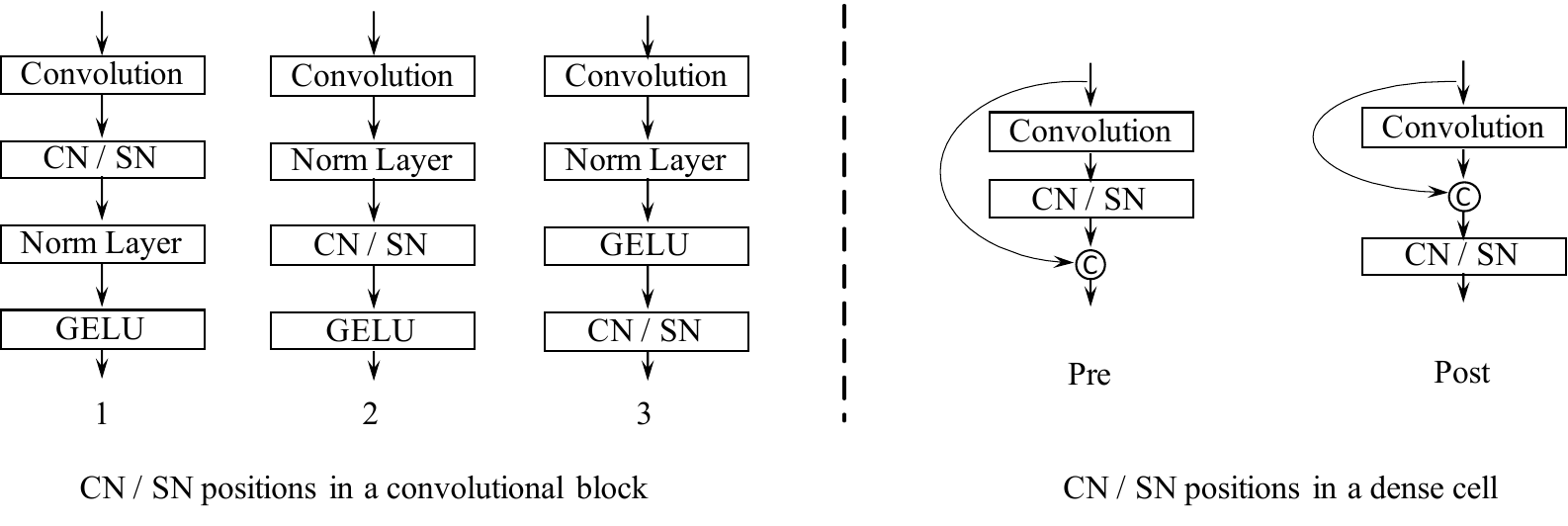}
\caption{Illustration of CN / SN positions in a AllConvNet block, and a dense cell of DenseNet. For a AllConvNet block, we name the position after convolution as \textit{1}, after normalization as \textit{2}, and after GELU as \textit{3}. Moreover, we investigate two locations in a dense cell, where we label the position before feature concatenation as \textit{Pre}, and after concatenation as \textit{Post}.}
\label{fig:sn_cn_position_allconv_dense}
\end{figure*}

\begin{figure*}[t]
\centering
\includegraphics[width=.9\linewidth]{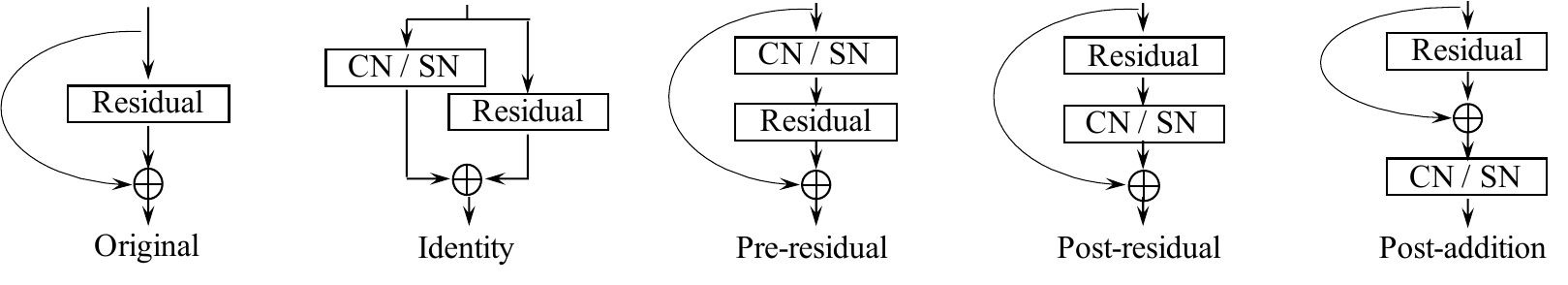}
\caption{Illustration of CN / SN positions in a residual module. We explore four positions: identity, pre-residual, post-residual, and post-addition.}
\label{fig:residual-positions}
\vspace{-1.5em}
\end{figure*}


\subsection{Ablation Study on CIFAR}
{\bf CN variants}. CN can be in 1-instance or 2-instance mode with four cropping options. According to Table \ref{tb:exp-1ins-2ins}, the 2-instance mode constantly gets lower errors than the 1-instance. Furthermore, cropping can help decrease the error since it can encourage style augmentation diversity. Note that style cropping may not always be superior. We will give a more detailed study on cropping choices later.

\begin{table}[h]
\begin{center}
\caption{1-instance CN {\it v.s.} 2-instance CN. We report the mCEs of WideResNet-40-2 on CIFAR-100-C.  The 2-instance mode consistently obtains lower errors than the 1-instance one. Moreover, proper cropping can further help decrease the errors. }\label{tb:exp-1ins-2ins}\vspace{-1ex}
\setlength\tabcolsep{0.5pt}
\begin{tabular}{l|cccc|cccc}
\toprule
 & \multicolumn{4}{c}{1-instance} & \multicolumn{4}{c}{2-instance}\\
\hline
Crop & Neither & Content & Style & Both & Neither & Content & Style & Both \\
\hline
mCE & 50.5 & 50.6 & 50.6 & 49.6 & 48.8 & 49.0 & {\bf 47.9} & 48.5\\
\bottomrule
\end{tabular}
\end{center}
\vspace{-1.5em}
\end{table}

{\bf Blocks choices for CN and SN}. Given both CN and SN placed at the post-addition location of a residual cell in WideResNet-40-2, we study which blocks in a network should build on the modified cells. No residual cell is required if applying CN and SN to the image space. According to Table \ref{tb:ablation-stage}, They both perform the best on CIFAR-100-C when plugged into all blocks.

\begin{table}[h]
\begin{center}
\caption{Exploration of Block choices for CN and SN. We compare the mCEs (\%) when applying CN and SN to image space or different blocks in WideResNet-40-2. Using them in all blocks gives the lowest errors on CIFAR-100-C.}\label{tb:ablation-stage}\vspace{-1ex}
\setlength\tabcolsep{3pt}
\begin{tabular}{lcccccc}
\toprule
Stages & N/A & Image & Block 1 & Block 2 & Block 3 & All\\
\hline
CN & 53.3 & 54.3 & 52.2 & 51.2 & 51.5 & {\bf 48.8}\\
\hline
SN & 53.3 & 52.9 & 48.9 & 52.2 & 51.3 & {\bf 47.4}\\

\bottomrule
\end{tabular}
\end{center}
\vspace{-1em}
\end{table}


{\bf Order of CN and SN}. In this experiment, we study two orders: SN$\rightarrow$CN and CN$\rightarrow$SN when plugging them into the post-addition place in all residual cells of WideResNet-40-2. Table \ref{tb:sncn-order} shows very close mCEs, indicating their order has little influence on the robustness performance.

\begin{table}[h]
\begin{center}
\caption{Order study of CN and SN. They both locate at the post-addition position in each residual cell of WideResNet-40-2, and we report the mCE on CIFAR-100-C.}\label{tb:sncn-order}\vspace{-1ex}
\begin{tabular}{lcccccc}
\toprule
Order & SN$\rightarrow$CN & CN$\rightarrow$SN\\
\hline
mCE(\%) & 46.9 & 46.6\\
\bottomrule
\end{tabular}
\end{center}
\vspace{-1.5em}
\end{table}


{\bf SN {\it v.s.} SE}. Although SN shares a similar attention mechanism with SE, SN obtains much lower corruption error than SE, according to Table \ref{tb:exp-sn-vs-se}. SN recalibrates the feature map statistics, suppressing unseen styles in the test data with distribution shifts, whereas SE, modeling the interdependence between feature channels, may not help generalization robustness.

\begin{table}[h]
\begin{center}
\caption{SN vs. SE. SN obtains much lower mCE than SE when they are applied to WideResNet-40-2 and CIFAR-100-C. We place SN in the post-addition location.}\label{tb:exp-sn-vs-se}\vspace{-1ex}
\setlength\tabcolsep{4pt}
\begin{tabular}{lcccc}
\toprule
& Basic & SE & SE(post-addition) & SN \\
\hline
mCE(\%) & 53.3 & 52.3 & 51.0 & {\bf 47.4}\\
\bottomrule
\end{tabular}
\end{center}
\vspace{-1em}
\end{table}

{\bf Modular positions}. Here we investigate the positions of CN and SN inside network cells. We give a comprehensive study on different cells and measure the performance on both CIFAR-10-C and CIFAR-100-C. Specifically, we conduct experiments using four backbones: AllConvNet, DenseNet, WIdeResNet and ResNeXt, consisting of three types of cells: naive convolutional cell, dense cell, and residual module, illustrated in Figures \ref{fig:sn_cn_position_allconv_dense} and \ref{fig:residual-positions}. According to Tables \ref{tb:sn-positions-cifar} and \ref{tb:cn-positions-cifar}, CN has stable best positions across the two datasets, while SN's optimal positions are different for CIFAR-10-C and CIFAR-100-C.

\begin{table}[h]
\begin{center}
\caption{Evaluation of {\bf SN} modular positions for AllConvNet, DenseNet, WIdeResNet and ResNeXt. The impacts of different positions are measured by mCE on both CIFAR-10-C ({\bf Top}) and CIFAR-100-C ({\bf Bottom}). Note that the four backbones have three types of cells whose positions are illustrated in Figures \ref{fig:sn_cn_position_allconv_dense} and \ref{fig:residual-positions}. }\label{tb:sn-positions-cifar}\vspace{-1ex}
\begin{tabular}{l|c|c|c|c}
\toprule
    \multicolumn{5}{c}{SN on CIFAR-10-C}\\
    \hline
     Position   &   1   &   2   &   3   & - \\
     AllConvNet &\textbf{24.0} & 26.4 & 25.6 &-\\
     \hline
     Position   & Pre & Post & -  &- \\
     DenseNet   &23.4 & \textbf{22.0}    &  - & -\\
     \hline
     Position   & Residual  & Post  & Pre   &Identity\\
     WideResNet & 22.7     &21.3  &\textbf{20.8} &22.3 \\
     \hline
     Position   & Residual  & Post  & Pre   &Identity   \\
     ResNeXt    &21.9  &24.8 &\textbf{21.5}   &22.0 \\
     \toprule
     \multicolumn{5}{c}{SN on CIFAR-100-C}\\
     \hline
     Position   &   1   &   2   &   3   & - \\
     AllConvNet &\textbf{50.3} & 51.6 & 51.0 & -\\
     \hline
     Position   & Pre & Post & - & -\\
     DenseNet   &\textbf{53.9} & 54.7     & - & -\\
     \hline
     Position   & Residual  & Post  & Pre   &Identity\\
     WideResNet & 49.3     &\textbf{47.4} &49.8 &48.4 \\
     \hline
     Position   & Residual  & Post  & Pre   &Identity   \\
     ResNeXt    &\textbf{47.6} &49.0 &50.9   &50.4 \\
     
\bottomrule
\end{tabular}
\end{center}
\vspace{-2em}
\end{table}

\begin{table}[h]
\begin{center}
\caption{Evaluation of {\bf CN} positions in the cells of four backbones. We measure the position influence by mCE on CIFAR-10-C ({\bf Top}) and CIFAR-100-C ({\bf Bottom}). The position choices in the naive convolution cell, dense cell, and residual module are shown in Figures \ref{fig:sn_cn_position_allconv_dense} and \ref{fig:residual-positions}.  }\label{tb:cn-positions-cifar}\vspace{-1ex}
\begin{tabular}{l|c|c|c|c}
\toprule
\multicolumn{5}{c}{CN on CIFAR-10-C}\\
    \hline
     Position   &   1   &   2   &   3   & - \\
     AllConvNet &\textbf{26.0} & 26.3 & 26.8 & -\\
     \hline
     Position   &  Pre & Post &  - & -\\
     DenseNet   &\textbf{24.7} & 29.2    & - & - \\
     \hline
     Position   & Residual  & Post  & Pre   &Identity\\
     WideResNet & 25.2     &\textbf{21.6}  &24.9 &23.3 \\
     \hline
     Position   & Residual  & Post  & Pre   &Identity   \\
     ResNeXt    &26.7  &\textbf{22.4} &23.8   &26.9 \\
     \toprule
     \multicolumn{5}{c}{CN on CIFAR-100-C}\\
     \hline
     Position   &   1   &   2   &   3   & - \\
     AllConvNet &\textbf{52.2} & 52.5 & 52.7 & - \\
     \hline
     Position   &  Pre & Post & - &- \\
     DenseNet   &\textbf{55.4} & 57.6    & - &-\\
     \hline
     Position   & Residual  & Post  & Pre   &Identity\\
     WideResNet & 52.1     &\textbf{48.8}  &51.7 &50.3 \\
     \hline
     Position   & Residual  & Post  & Pre   &Identity   \\
     ResNeXt    &51.5  &\textbf{47.0} &47.9  &50.2 \\
\bottomrule
\end{tabular}
\end{center}
\vspace{-2em}
\end{table}

\begin{table}[h]
\begin{center}
\caption{Study of CN cropping choices. We evaluate four cropping choices: neither, content, style, and both when jointly using CN with SN in four backbones. The performance is measured by mCE on both CIFAR-10-C ({\bf Top}) and CIFAR-100-C ({\bf Bottom}). We put the modular positions next to the backbone names. }\label{tb:exp-cropping-sncn}\vspace{-1ex}
\setlength\tabcolsep{3pt}
\begin{tabular}{l|cccc}
\toprule
\multicolumn{5}{c}{CNSN with cropping on CIFAR-10-C}\\
    \hline
     Backbone   & Neither   & Content   &Style  &Both\\
\midrule
     AllConvNet, 1 &19.0 & 20.3 & \textbf{18.8} &20.3\\
     \hline
     DenseNet, Conv1 Pre   &18.8 & \textbf{18.2}     &18.7  &18.8\\
     \hline
     WideResNet, Post & 17.9     &18.0 &\textbf{16.8} &17.5 \\
     \hline
     ResNeXt, Post  &\textbf{17.7} &18.5 &18.4   &18.6 \\
     \toprule
     \multicolumn{5}{c}{CNSN with cropping on CIFAR-100-C}\\
     \hline
     Backbone   & Neither   & Content   &Style  &Both\\
\midrule
     AllConvNet, 1 &44.2 & 46.9 & \textbf{43.9} &46.1\\
     \hline
     DenseNet, Conv1 Pre   &51.4 & 49.4     &49.1   &\textbf{49.0}\\
     \hline
     WideResNet, Post & 46.6     &45.1 &45.8 &\textbf{44.5} \\
     \hline
     ResNeXt, Post  &\textbf{41.0} &44.9 &43.0   &46.5 \\
     
\bottomrule
\end{tabular}
\end{center}
\vspace{-2em}
\end{table}

\begin{table*}[t]
\begin{center}
\caption{Incremental ablation study for CN, SN, cropping, consistency regularization and AugMix. We report the mCEs of four backbones on both CIFAR-10-C ({\bf Top}) and CIFAR-100-C ({\bf Bottom}). The modular position and cropping choice are also given in each row.}\label{tb:exp-incremental-ablation}
\begin{tabular}{l|c|c|c|c|c|c|c}
\toprule
    \multicolumn{8}{c}{CIFAR-10-C}\\
    \hline
     Backbone   & Basic   & CN   &SN  &CNSN &CNSN+Crop & AugMix & CNSN+Crop\\
                &         &      &    &+Crop  &+Consistency &      &+AugMix   \\
\midrule
     AllConvNet, 1, style &30.8 & 26.0 & 24.0 &18.8  &17.2  &15.0 &\textbf{11.8}\\
     \hline
     DenseNet, Conv1 Pre, both   &30.7 & 24.7 & 22.0  &18.8 &18.5 &12.7 & \textbf{10.4} \\
     \hline
     WideResNet, Post, both & 26.9  &21.6  &20.8  &17.5    &16.9  &11.2  &\textbf{9.9} \\
     \hline
     ResNeXt, Post, neither  &27.5 &22.4  &21.5   &17.7 &15.7  &10.9   &\textbf{9.1} \\
     \toprule
     \multicolumn{8}{c}{CIFAR-100-C}\\
     \hline
     Backbone   & Basic   & CN   &SN  &CNSN &CNSN+Crop & AugMix & CNSN+Crop\\
                &         &      &    &+Crop  &+Consistency &      &+AugMix   \\
\midrule
     AllConvNet, 1, style &56.4 & 52.2 & 50.3 &43.9  &42.8  &42.7  &\textbf{36.8}\\
     \hline
     DenseNet, Conv1 Pre, both   &59.3  &55.4  & 53.9 &49.0 &48.5 &39.6 & \textbf{37.0} \\
     \hline
     WideResNet, Post, both & 53.30  &48.8  &47.4  &44.5    &43.7  &35.9  &\textbf{33.4} \\
     \hline
     ResNeXt, Post, neither  &53.4 &47.0 &47.6    &41.0  &40.8  &34.9   &\textbf{30.8} \\
\bottomrule
\end{tabular}
\end{center}
\vspace{-1em}
\end{table*}

{\bf Cropping Choices for CN}. Cropping enables diverse statistics transfer between feature maps. Here we study four cropping choices: neither (no cropping), style, content, and both. In Table \ref{tb:exp-cropping-sncn}, we can find the best cropping choice may change over backbones and datasets.




{\bf Incremental ablation study}. CN and SN are generic normalization techniques to improve the generalization robustness under distribution shifts. They are orthogonal to each other, and other methods such as the consistency regularization \cite{hendrycks2020augmix} and AugMix \cite{hendrycks2020augmix}. According to Table \ref{tb:exp-incremental-ablation}, they can lower the corruption error both separately and jointly. On top of them, proper cropping, consistency regularization, and domain-specific AugMix can further advance the generalization robustness. 



\subsection{Ablation Study on ImageNet}
{\bf CN {\it v.s.} Stylized-ImageNet}. We also compare CN to Stylized-ImageNet, which transfers styles from external datasets to perform style augmentation. Stylized-ImageNet finetunes a pre-trained ResNet-50 for 45 epochs with double data (stylized and original ImageNets) in each epoch. To compare CrossNorm with Stylized-ImageNet, we perform the finetuning for 90 epochs using only the original ImageNet. In Table \ref{tb:exp-cn-sin}, although Stylized-ImageNet has 2\% lower corruption error than CN, its clean error is 3.8\% higher, because the external styles in Stylized-ImageNet cause large distribution shifts, impairing its clean generalization. In contrast, CN can use more consistent yet diverse internal styles to decrease both corruption and clean errors. 

\begin{table}[h]
\begin{center}
\caption{Comparison of Stylized-ImageNet and CN. Following the Stylized-ImageNet setup, we finetune a pre-trained ResNet50 model 90 epochs on ImageNet. Compared with SIN, CN holds a better balance between clean and corruption errors. }\label{tb:exp-cn-sin}
\setlength\tabcolsep{4pt}
\begin{tabular}{lccc}
\toprule
& Basic & Stylized-ImageNet & CN \\
\hline
Clean error (\%) & 23.9 & 27.2 & {\bf 23.4}\\
mCE(\%) & 80.6 & {\bf 73.3} & 75.3\\
\bottomrule
\end{tabular}
\end{center}
\vspace{-1em}
\end{table}

{\bf CN and SN locations}. Moreover, in Table \ref{tb:exp-sncn-pos-imagenet}, we also investigate the CN and SN locations in a residual module using ImageNet and ResNet50. Similar to the CIFAR results, the post-addition position performs the best for corruption robustness.

\begin{table}[h]
\begin{center}
\caption{Investigation of CN ({\bf Top}) and SN ({\bf Bottom}) positions in a residual module of ResNet50 trained 90 epochs on ImageNet. 
}\label{tb:exp-sncn-pos-imagenet}
\setlength\tabcolsep{2pt}
\begin{tabular}{lcccc}
\toprule
\multicolumn{5}{c}{CN modular positions}\\
\hline
\multirow{2}{*}{Position} & \multirow{2}{*}{Identity} & Pre- & Post- & Post-\\
& & Residual & Residual & Addition\\
\hline
Clean error (\%) & 25.2 & {\bf 23.4} & 23.5 & {\bf 23.4}\\
mCE(\%) & 78.2 & 75.8 & 77.5 & {\bf 75.3}\\
\toprule
\multicolumn{5}{c}{SN modular positions}\\
\hline
\multirow{2}{*}{Position} & \multirow{2}{*}{Identity} & Pre- & Post- & Post-\\
& & Residual & Residual & Addition\\
\hline
Clean error (\%) & 24.0 & {\bf 23.0} & 23.2 & 23.7\\
mCE(\%) & 75.5 & 75.8 & 74.8 & {\bf 73.8}\\
\bottomrule
\end{tabular}
\end{center}
\vspace{-1em}
\end{table}

{\bf Ablation study with IBN}.
Table \ref{tb:imagenet-ablation} reports the results of applying CN or SN with IBN. We can observe that they can cooperate to improve the corruption robustness of ResNet50. Moreover, integrating CN, SN, IBN, and AugMix can bring the lowest corruption error. This shows CN and SN's advantage that they are generic to boost other state-of-the-art methods. 

\begin{table*}[t]
\begin{center}
\caption{Ablation study of IBN, CN, SN, consistency regularization(CR), and AugMix(AM) on ImageNet-C with ResNet50. IBN, initially designed for domain generalization, can also decrease mCE. Both CN and SN can further lower the error based on IBN. Combining them with AM gives the best robustness performance. }\label{tb:imagenet-ablation}
\setlength\tabcolsep{1.5pt}
\begin{tabular}{l|c|c|c|c|c|c|c|c|c}
\toprule
& ResNet50 & \multicolumn{4}{|c}{ResNet50+IBN(a)} & \multicolumn{4}{|c}{ResNet50+IBN(b)}\\
\hline
& Basic & Basic & +CN & +CN+CR & +CN+AM & Basic & +SN & +SN+AM & +CNSN+AM\\
\hline
Clean err(\%) & 23.9 & 23.2 & 23.1 & 22.6 & 22.5 & 24.0 & 23.5 & {\bf 22.3} & {\bf 22.3}\\
mCE(\%) & 80.6 & 75.1 & 73.2 & 73.6 & 66.4 & 74.1 & 72.6 & 64.1 & {\bf 62.8}\\
\bottomrule
\end{tabular}
\end{center}
\vspace{-2em}
\end{table*}

\begin{figure*}[t]
\centering
\includegraphics[width=1\linewidth]{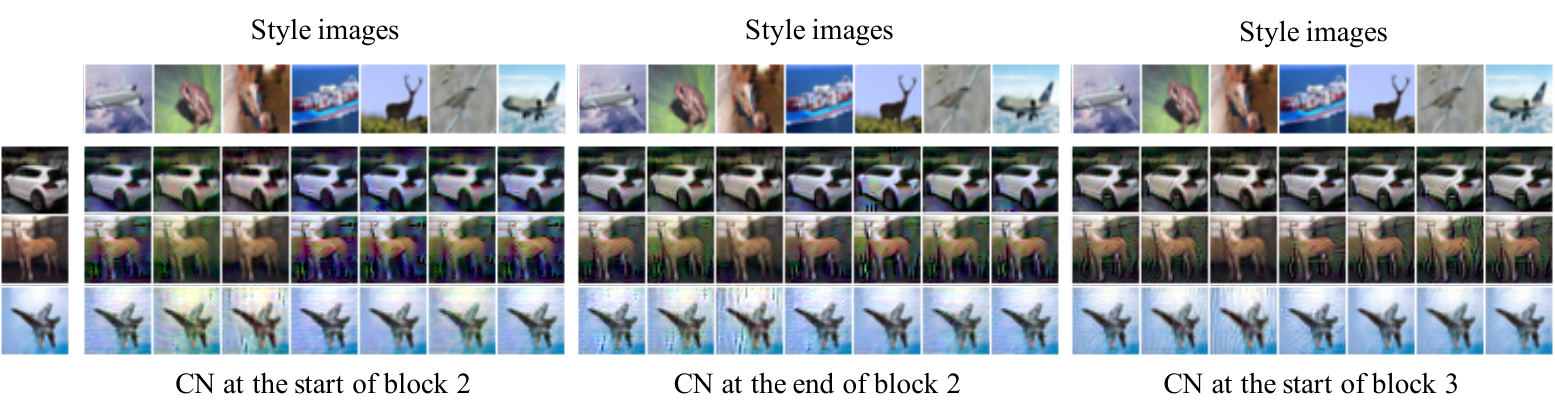}
\caption{CN visualization at the head ({\bf Left}), the tail of ({\bf Middle}) block 2 and the start of block 3 ({\bf Right}) in a WideResNet-40-2. Both the content ({\bf Row}) and style ({\bf Column}) images are from CIFAR-10. Compared to CNs in block 1, shown in Figure 5 of the paper, the CNs in blocks 2 and 3 have weaker style transfer effects because the statistics in high-level feature maps may contain less low-level visual information.}
\label{fig:cn-vis-2}
\end{figure*}

\begin{figure}[t]
\centering
\includegraphics[width=\linewidth]{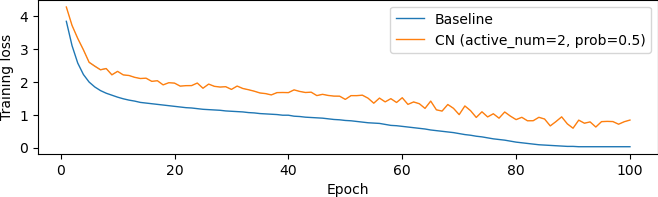}
\caption{Training loss curves. The CN curve is an average of 5 runs. Compared to the baseline, CN produces larger training losses, reducing the overfitting to training data.}
\label{fig:train-curve}
\vspace{-1em}
\end{figure}

\begin{figure*}[htb]
\centering
\includegraphics[width=1\linewidth]{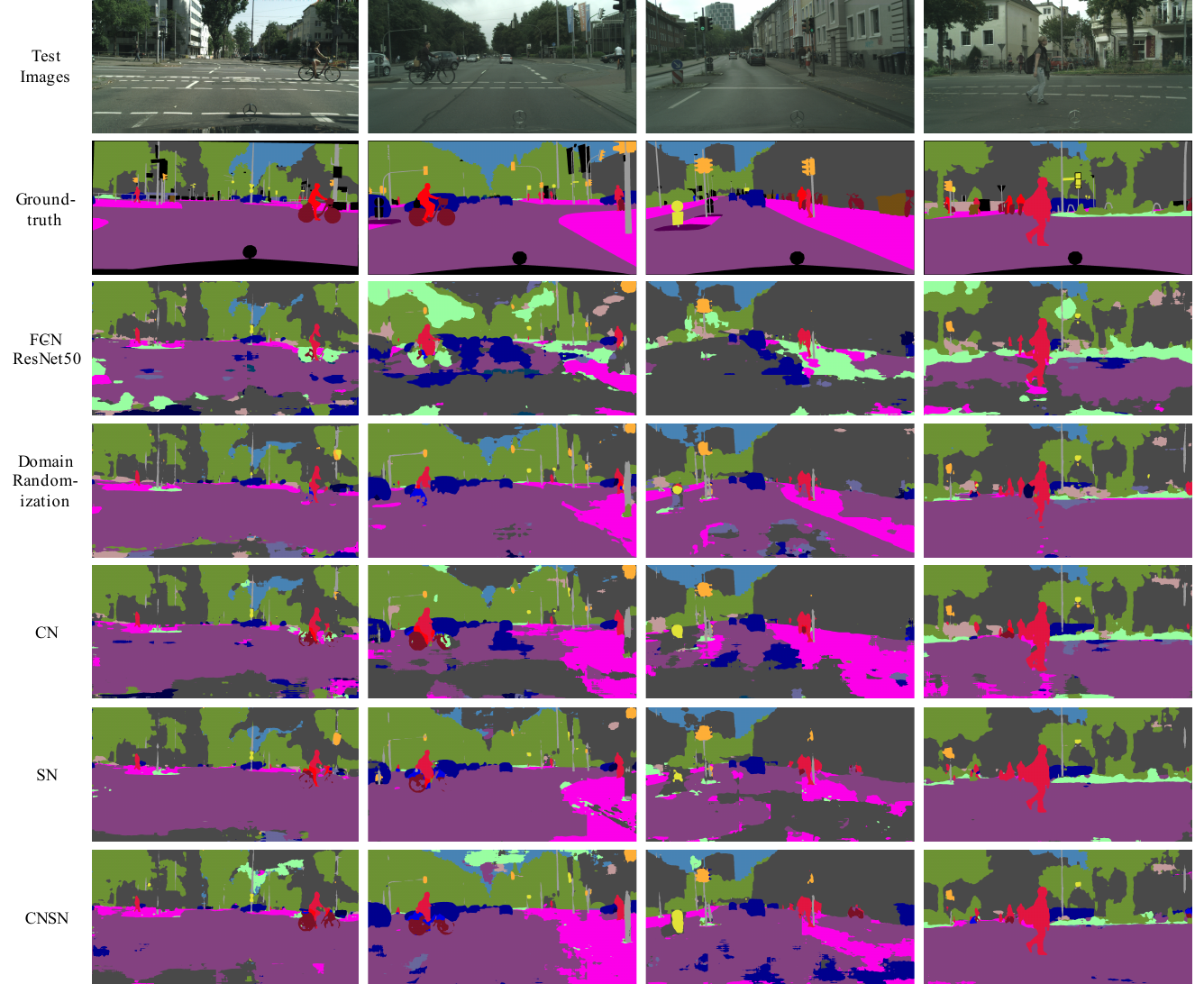}
\caption{A visualization of segmentation results. All models are trained on GTA5 and tested on Cityscapes. We show the results of four images ({\bf First row}) with corresponding ground-truth annotations ({\bf Second row}). Due to the domain gap between GTA5 and Cityscapes, the baseline FCN-ResNet50 predicts poor segmentation masks ({\bf Third row}). Based on this baseline model, we compare the qualitative results of domain randomization
({\bf Forth row}), CN ({\bf Fifth row}), SN ({\bf Sixth row}), and CNSN ({\bf Seventh row}).  Domain randomization can largely improve the baseline performance, but it requires multiple style transfer networks and additional style datasets to conduct style augmentation. With little overhead, our CN and SN can also make clear improvements on the baseline line. Combining CN and SN produces the best visualization results, demonstrating their complementary property. Additionally, CN performs better than domain randomization on predicting small masks of human, vehicle, and object. Yet, domain randomization has fewer fragments than CN for large masks of the flat, such as road and sidewalk. One feasible explanation is that CN can generate some local artifacts, shown in the third column in Figure \ref{fig:seg-images-cn-vis}, that are more likely to affect predicting large masks.}
\label{fig:seg-results-vis}
\end{figure*}

\begin{figure*}[t]
\centering
\includegraphics[width=1\linewidth]{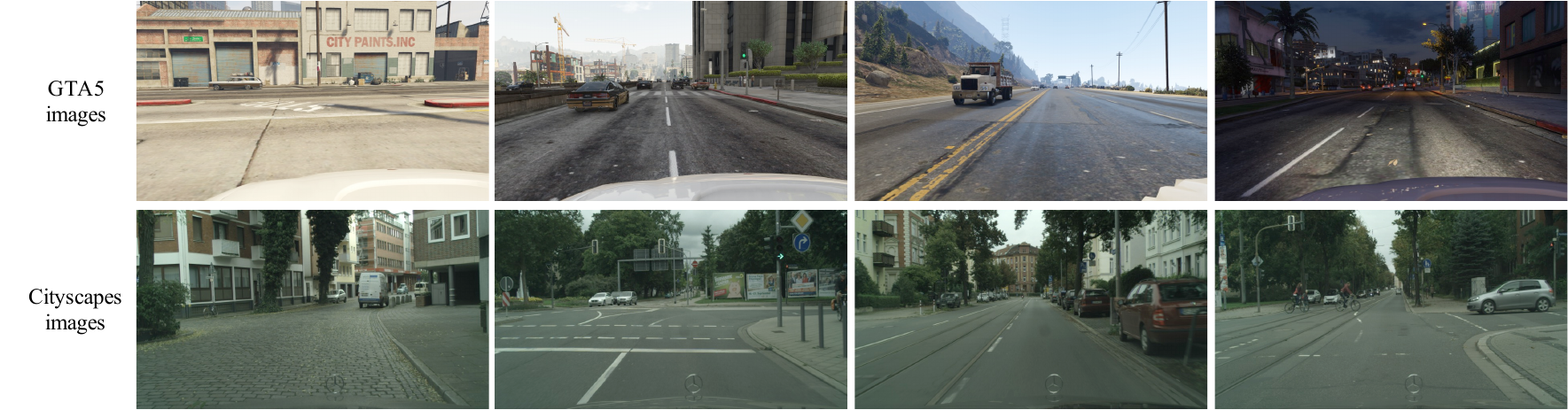}
\caption{Some examples in segmentation datasets GTA5 ({\bf Top}) and Cityscapes ({\bf Bottom}). We can find that the two datasets have a clear domain gap.}
\label{fig:seg-images-vis}
\end{figure*}


\begin{figure*}[t]
\centering
\includegraphics[width=1\linewidth]{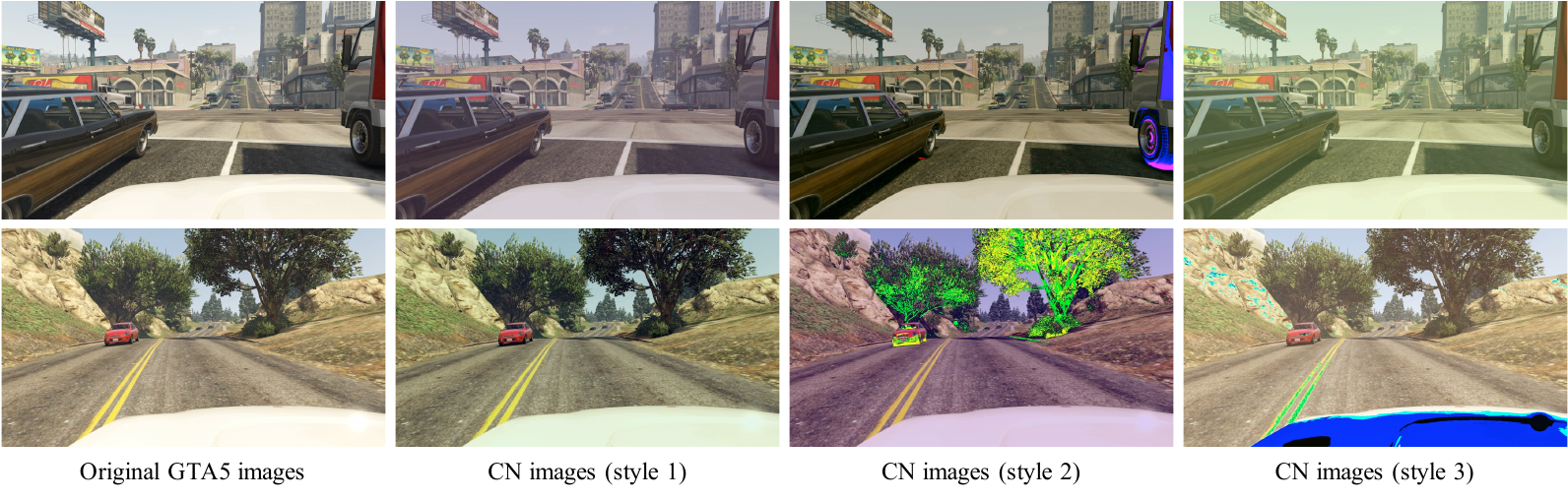}
\caption{A visualization of applying CN to GTA5 images with style cropping. The two images on the first column are the original images in the GTA5 dataset. We apply CN to these two images with three different style cropping, resulting in images in the last three columns. By calculating style statistics from random cropping, CN can perform diverse style augmentation. Note that we use CN in both image and feature levels, and here we only visualize in the image level for simplicity.}
\label{fig:seg-images-cn-vis}
\end{figure*}

\begin{figure*}[h!]
\centering
\includegraphics[width=.6\linewidth]{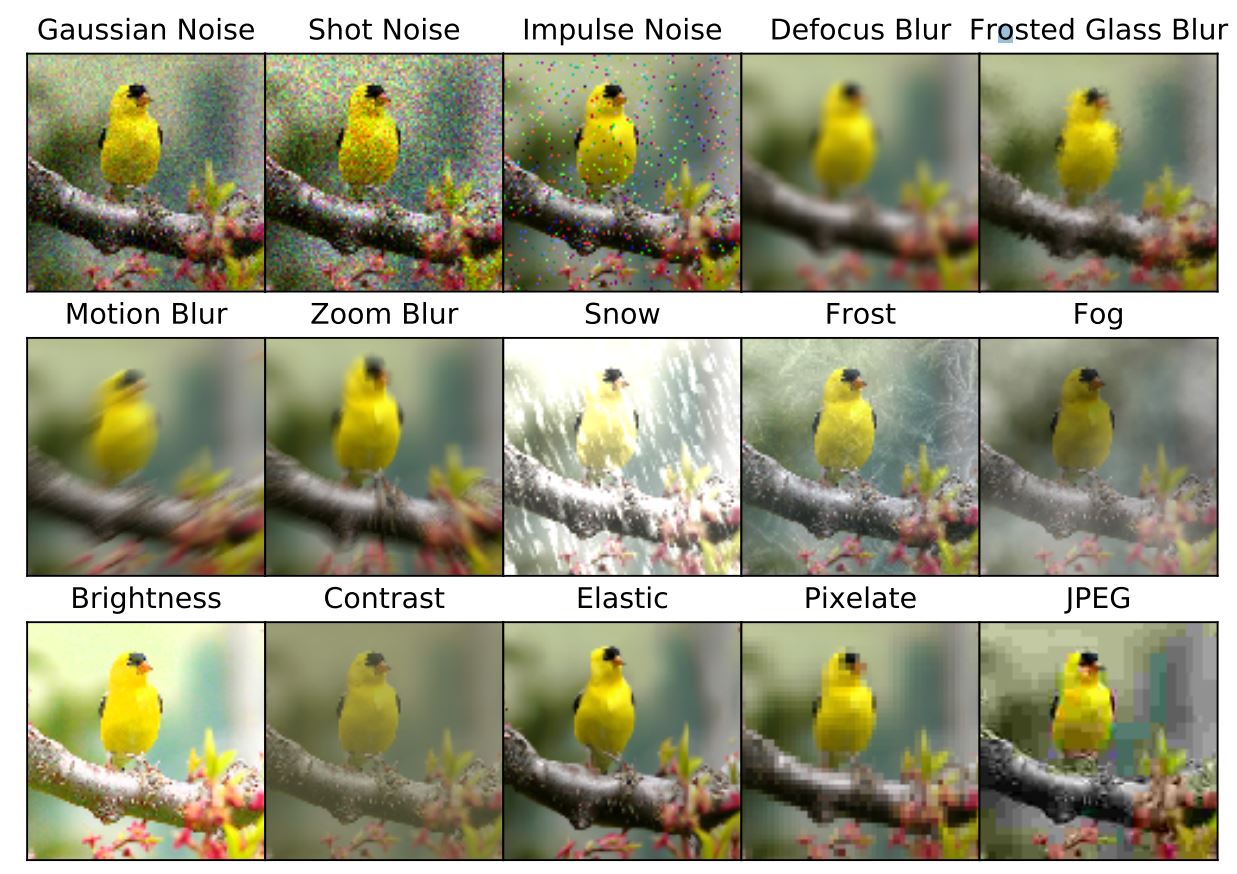}
\caption{A demostration of corrupted images in ImageNet-C dataset \cite{hendrycks2019benchmarking}. Fifteen types of algorithmically generated corruptions from noise, blur, weather, and digital categories are applied to images to create the corrupted dataset.}
\label{fig:imagenet-c-vis}
\end{figure*}

\subsection{CN Effect on Training Loss}
Similar to other data augmentation
methods, CN has larger training losses than the baseline.
This is a normal phenomenon of reducing overfitting instead of making training unstable. Our 5 runs of CN with
WideResNet-40-2 on CIFAR-100 indicate that the mean
training loss decreases from 4.3 to 0.8, according to Figure \ref{fig:train-curve}. The
loss std per epoch is at the order of 1e-3.

\subsection{Visualization Continued}
{\bf Visualization setup}. As it is nontrivial to visualize the effect of CN and SN directly in feature space, we perform the visualization with the technique: understanding deep image representations by inverting them \cite{mahendran2015understanding}. The goal is to find an image whose feature representation best matches the given one. The search is done automatically by a SGD optimizer with learning rate 1e4, momentum 0.9, and 200 iterations. The learning rate is divided by 10 every 40 iterations. During the optimization, the network is in its evaluation mode with its parameters fixed. In the experiment, we use WideResNet-40-2 and images from CIFAR-10. In visualizing CN, we use the training images and a model trained for 1 epoch. The SN visualization uses test images and a well-trained model. We use different settings for them because CN is for training, while SN works in testing.


{\bf More visualization results}. Here we provide more visualization results in addition to those in the paper. First, Figure \ref{fig:cn-vis-2}, extending Figure 5 in the paper, shows more CN visualizations in blocks 2 and 3 in WideResNet-40-2. Second, we visualize the segmentation results on four Cityscapes' images for models trained on GTA5 in Figure \ref{fig:seg-results-vis}, where we compare baseline FCN-ResNet50, domain randomization, CN, SN, and CNSN. Moreover, Figure \ref{fig:seg-images-vis} shows four synthetic images from GTA5 and four realistic ones from Cityscapes, and Figure \ref{fig:seg-images-cn-vis} visualizes the effect of applying CN to two GTA5 images. Finally, Figure \ref{fig:imagenet-c-vis} gives an illustration of 15 corruptions used in evaluating the corruption robustness on CIFAR and ImageNet.

\end{document}